# Study of Phonemes Confusions in Hierarchical Automatic Phoneme Recognition System

*Rimah Amami, Noureddine Ellouze
*Department of Electrical Engineering*
*National Engineering School of Tunis, Tunisia*
* rimah.amami@yahoo.fr

**Abstract**
*In this paper, we have analyzed the impact of confusions on the robustness of phoneme recognitions system. The confusions are detected at the pronunciation and the confusions matrices of the phoneme recognizer. The confusions show that some similarities between phonemes at the pronunciation affect significantly the recognition rates. This paper proposes to understand those confusions in order to improve the performance of the phoneme recognition system by isolating the problematic phonemes. Confusion analysis leads to build a new hierarchical recognizer using new phoneme distribution and the information from the confusion matrices.*
*This new hierarchical phoneme recognition system shows significant improvements* of the recognition rates on TIMIT database.

## 1. Introduction

The impact of the phoneme confusions through speech recognitions system is a facet that remains to a large extent unexplored. The precision of automatic speech recognition (ASR) is still behind the human speech recognition since it is so robust to speech variations i.e. the same speaker may use different speaking styles for the same sentence, accent, speaking rate, etc. [1] [2] [3].Those variations can cause severely the degradation of the performance of the phoneme recognition system. Besides, the performance of the classifiers depends mainly on the type of used datasets which may misled the selected classifier and it will not yield the best performance.

Thus, the aim of this study is to analyze the impact of phoneme confusions on the recognition rates and to investigate the solution to reduce those confusions. Indeed, this confusion analysis should help to identify sources errors and to improve the performance of the phoneme recognition system. Moreover, the confusion analysis will serve to construct our new hierarchical phoneme recognition system which is isolating and limiting the phoneme confusions.
In this study we used the most common features in speech recognition system MFCC [4] [5] and the classifier SVM [6] [7].
The rest of the paper is organized as follows: Section 2 presents the analysis of phoneme confusions into TIMIT database using the confusions matrices produced by the phoneme recognizer and the pronunciation confusions. Section 3 describes the architecture of the new hierarchical system dealing with the phonemes confusions. The effectiveness of the proposed hierarchical system is demonstrated in the section 5 in terms of the recognition rates per phoneme. Finally, section 6 presents the conclusion and scope for future works.

## 2. Analysis of phonemes confusions

The speech database used for the confusion analysis is TIMIT [8]. It consists 6300 sentences where each, speaker have spoken 10 speech sentences. The TIMIT database is labeled using 60 phonetic labels. For our experiments, we used the phonemes from the dialect DR1 of TIMIT [9].
Traditionally, in the literature, the phonemes of TIMIT are organized into 7 sub-groups depending on the articulatory features of phonemes [10].
The first step of our experiments is to study the confusion. Thus, the analysis of the phoneme confusion is performed on hierarchical phoneme recognition system (called HS-TC) based on two levels: The first level is composed of all phonemes in aim to classify the phoneme to one of the 7 sub-group into which it belongs. The second level consists of 7 sub-phonetic groups where the goal is to recognize the tested phoneme. It must be pointed out that the phonemes of the same sound are grouped under the same class.

Table 1. The sub-phonemes groups

| Sub-Phonemes | Units |
|---|---|
| Vowels | [/aa/ /ae/ /ah/ /ao/ /aw/ /ax/ /ax-h/ /axr/ /ay/ /eh/ /er/ /ey/ /ih/ /ix/ /iy/ /ow/ /oy/ /uh/ /uw/ /ux/] |
| Semivowels | [/l/ /r/ /w/ /y/ /hh/ /hv/ /el/] |
| Stops | [/b/ /d/ /g/ /p/ /t/ /k/ /dx/ /q/ /bcl/ /dcl/ /gcl/ /pcl/ /tcl/ /kcl/] |
| Other stops | [/pau/ /epi/ /h#/] |
| Nasals | [/m/ /n/ /ng/ /em/ /en/ /nx/] |
| Affricates | [/ch/ /jh/] |
| Fricatives | [/s/ /sh/ /z/ /zh/ /f/ /th/ /v/ /dh/] |

At this stage of experiments, we will compare the pronunciation of each phoneme spoken with the official dictionary pronunciation given by TIMIT [9] and, also, the confusions matrices obtained by performing phoneme classification using SVM. Actually, every word in English has different pronunciations resulting from many factors such as the similarities in articulatory features between phonemes, rhythm of articulation of the speaker, and dialect. Moreover, one phoneme in English may have various possible pronunciations. Thus, for the pronunciation of each phoneme in TIMIT, there are possible pronunciation confusions.

The analysis of that phoneme confusions matrices show that it exists a consistent problematic similarity at the pronunciation of phonemes from the same type (same group of sound).

For example, in TIMIT dictionary, the word she is interpreted as *sh iy*. Meanwhile, it exists several possible phonetic pronunciations in the database of this word including *sh ix, sh ih*, and *s uw,* etc [11]. However, the most correct pronunciation of the word she is the phonemes, sh iy and the others combinations are considered as errors. A list of the major confusions for each phoneme is displayed in the table 1. On the other hand, we believe that the analysis of confusion matrix may provide valuable information for detecting the complementary nature of information present in a set of classifiers that should be used to consolidate the classifier performance. For example, if the phoneme /ih/ is incorrectly classified, then it is more likely that the probability mass assigned to vowels such as /iy/ or /eh/ is higher.

Therefore, this information could be used to correct the output of the SVM classifier. As seen in table 2, it exists confusions of the vowels with other phonemes from the same class vowels and, also, the phonemes which belong to consonants group are confused with other phonemes from the same group, etc. For example, with s there several intensely confusions with *sh* and *z*. Those confusions between phonemes from the same sound group are due highly to the similar pronunciation of phonemes in English and lead to be easily confused.

The phonemes confusions as shown in table 2 are exploited to conceive a new hierarchical phoneme recognition system. It must be pointed out that the confusions matrices contain more confusions than the confusions contains in the official pronunciation dictionary of TIMIT.

**Table 2**. Major confusions from the pronunciation dictionary and from all confusions matrices of the hierarchical phoneme system based on SVM. Some minor confusions are eliminated

| Phonemes | Pronunciation confusions | Confusion matrix |
|---|---|---|
| iy | ix,ih | ey,ih,ix |
| ih | ix,iy,ax,eh | ix,iy,ax,eh,ey |
| eh | ih,ix | ix,ae,ah |
| ae | eh,ix | eh |
| ix | ih,ax,en,iy | ax,ih,ey,iy,eh,axr |
| ax | ix,ah,ih | ah,ix,ow,axr |
| uw | ux,ix,uh | ux,ax,axh |
| uh | ix,er,ax | ax,ix,eh,ah |
| ah | ax,ix | eh,ax,aa |
| ao | aa | aa |
| aa | ah,ao | ah,ao |
| er | axr,ax | axr,eh,ix |
| axr | er,ax,ix | ax,er,ix |
| ey | eh | ix,iy,ih |
| ay | aa | aa,eh |
| oy | ao,ow | ao,eh,ah |
| aw | aa | aa,ae,ah |
| ow | ax,uh | ax,ao,ah,aa |
| t | dx,q,d | d,k |
| k | - | t,d,tcl,p |
| q | - | dx,k,d |
| b | v | d,dx,q |
| d | dx,t | k,dx,t,tcl |
| g | - | k,dx,b |
| m | em | n |
| n | nx,en | m |
| ng | n | n |
| n | - | m |
| f | - | th,dh |
| th | dh,t | dh,f,s |
| v | f | dh,f |
| dh | th,d | th,f,v |
| z | s,zh | s,sh |
| zh | jh,z,sh,ch | sh,v,f |
| ch | sh | jh |
| r | axr,er | l,y |
| y | ix,ux | r,hv |
| em | m | m |
| en | ix,n | n |
| el | l | l,w |
| hh | hv | hv |

It concluded that the confusions from the dictionary matrices and from the confusions matrices of the phoneme recognizer based on SVM are slightly different. It can be seen that there is large similarities between phonemes which lead to reduce the performance of the phoneme recognizer. The most important confusions are identified with the consonants, vowels and semi-vowels. Those groups of phonemes show more confusions in the confusions matrices than in the pronunciation matrices. Thus, through the confusions matrices, we see that the confusions have greatest impact on phonemes which are likely to be mispronounced.

## 3. Architecture of the proposed hierarchical phoneme recognition system

In the previous section, we have analyzed the phoneme confusions for a hierarchical phoneme recognition system based on a traditional distribution of phonemes (all phonemes of similar articulatory structure are grouped into the same category). Indeed, SVM classifier was unable to distinguish efficiently the phonemes that are pronounced very similar.

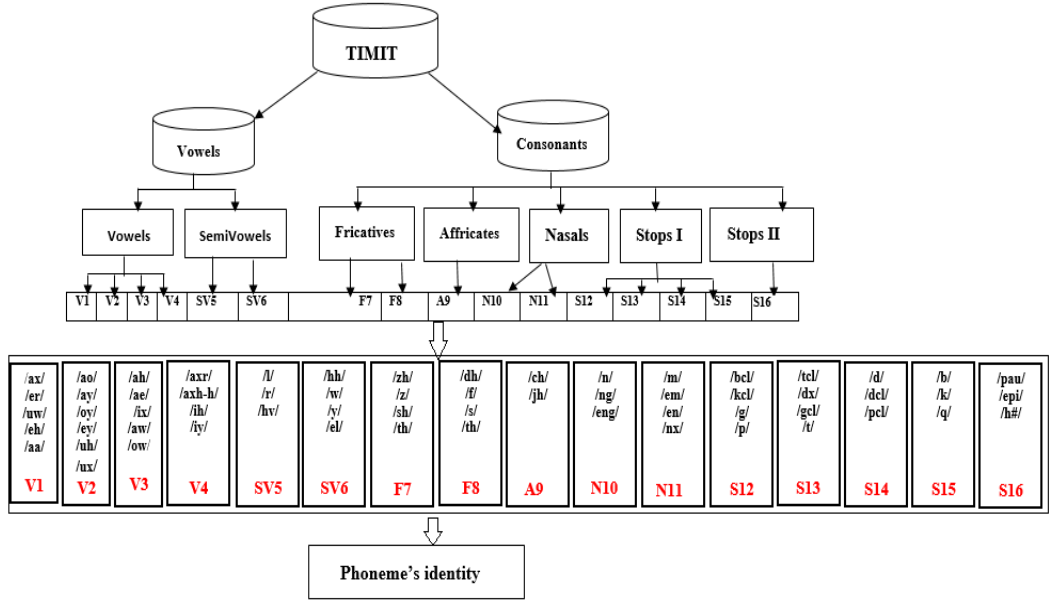

**Figure 1.** The architecture of the proposed hierarchical phoneme recognition system

For example, aa, ah, and ao are all intensely confused with each other. On the one hand, it is a fact that those phonemes are very similar that the distinction between them could be ignored. Nevertheless, in practice, the phoneme recognition system will consider this confusion as an error.

Thus, in order to improve the recognition rates, we proceed to the isolation of phonemes that are confused with each other (see table 2). In addition, the new structure led to bring up a new hierarchical phoneme recognition system (see figure 1).

The figure 1 presents the new hierarchical system (called HS-CO) with four levels; the first and second level allows recognizing the class of the phonetic sound (vowel or consonant). The third level is used to recognize the class of the phoneme with the new distribution that isolates confused phonemes with each other. Finally, the fourth level with which its new structure takes into account the most confused phonemes allow to better recognize the identity of the phoneme.

## 4. Experimental Conditions

Current phoneme recognition system is based on SVM classifier. This SVM based recognizer consists typically of 3 principal main modules: feature extraction (parameterization), training which consist of generating the SVM model, and then recognize the phoneme's identity. The figure 2 presents a brief description of our phoneme recognition system setup which is relevant for the confusion analysis. Concerning the parameterization module, two types of features are most standardly used in speech recognition systems: perceptually based linear predictive cepstral coefficients (PLP) and mel-frequency cepstral coefficients (MFCC). All our experiments were carried out just with these the last feature type MFCC.

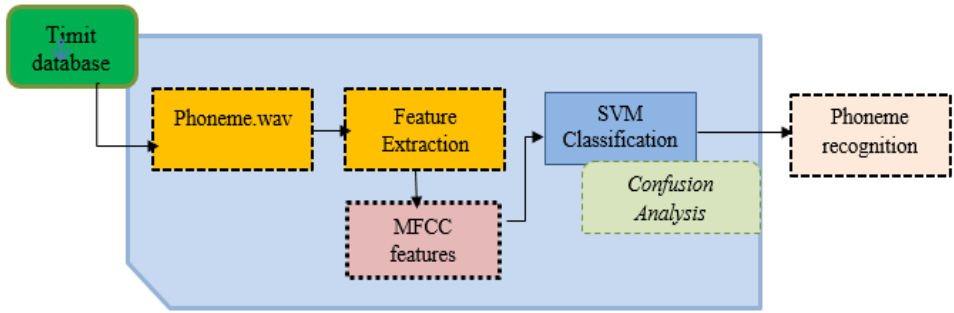

**Figure 2**. Block scheme of experimental setup

In fact, one of the fundamental denominator of all recognition system is the feature extraction since all of the information necessary to distinguish phonemes is preserved during the this stage. So, if important information is lost during the feature extraction, the performance of the following recognition stage will be inherently deteriorated which will affect the system outcome. Mel-frequency cepstral coefficients (MFCC) are derived from a type of cepstral representation. Davis and Mermelstein [12] were the first who introduced in 80's the MFCC concept for automatic speech recognition. The main idea of this algorithm consider that the MFCC are the cepstral coefficients calculated from the mel-frequency warped Fourier transform representation of the log magnitude spectrum.

The Delta and the Delta-Delta cepstral coefficients are an estimate of the time derivative of the MFCCs. In order to improve the performance of speech recognition system, an improved representation of speech spectrum can be obtained by extending the analysis to include the temporal cepstral derivative; both first (delta) and second (delta-delta) derivatives are applied. Those coefficients have shown a determinant capability to capture the transitional characteristics of the speech signal that can contribute to ameliorate the recognition task.

On the other hand, the particularity of MFCC is that those features use their own non-linear filter-bank which in both cases models the perceptual properties of human auditory system [13] [14] [15][16]. The MFCC of order 0 is not part of the feature vector that consists of the remaining 13 components as well as of the corresponding delta and acceleration coefficients. Thus a vector contains 39 components in total.

On the other hand, a brief introduction is needed to understand the classifier used by our phoneme recognition system.

Support Vector Machine (SVM) has been a very successful supervised algorithm for solving two-class and multi-class recognition problems. SVM is a method which is based on the Vapnic-Chervonenkis (VC) theory and the principle of structural risk minimization (SRM) [17][18] [19]. SVM approximates the solution to the minimization problem of SRM through a Quadratic Programming optimization. It aims to maximize the margin which is the distance from a separating hyperplane to the closest positive or negative sample between classes. A subset of training samples is chosen as support vectors.

Furthermore, the excellent learning performance of this method comes out from the fact that SVM apply a linear algorithm to the data in a high dimensional space. Thus, SVM can overcome dimensional space and overfitting problems in any learning system.

At this stage of the study, we supposed that we used MFCC 39-dimensional feature vectors, the kernel trick used is RBF and we set C = 10 and gamma= 0.027 (C is the regularization parameter and gamma is the kernel width).

## 5. Experimental Results

Author names and affiliations are to be centered beneath the title and printed in Times New Roman 11-point, non-boldface type. Multiple authors may be shown in a two or three-column format, with their affiliations below their respective names. Affiliations are centered below each author name, italicized, not bold. Include e-mail addresses if possible. Follow the author information by two blank lines before main text.

The recognition results are presented in this section when applying the SVM classifier, the MFCC features representation and the scheme as described above. We have built a model that deal at best with the phonemes exhibiting similarities into the same phonetic class having similar articulatory structure. Those phonemes confusions are causing errors and then, reducing the robustness of the phoneme recognizer. The table 3 presents a comparison of recognition rates with the hierarchical phoneme recognition system which don't take into account the phonemes confusions and the new hierarchical phoneme recognition based on new phoneme distribution which isolate phonemes similar to each other.

**Table 3.** Recognition results per phoneme with the hierarchical phoneme recognition system using traditional distribution (Traditional HS-TC) and the new hierarchical phoneme recognition system using the new distribution isolating the phonemes confusions (New HS-CO).

| Phoneme | HS-TC (Traditional) | HS-CO (New) |
|---|---|---|
| Aa | 54% | 59% |
| Ae | 65% | 66% |
| Ah | 27% | 43% |
| Ao | 52% | 67% |
| Aw | 9% | 15% |
| Ax | 54% | 63% |
| Axh | 35% | 55% |
| Axr | 51% | 58% |
| Ay | 60% | 68% |
| Eh | 43% | 61% |
| Er | 25% | 44% |
| Ey | 44% | 72% |
| Ih | 32% | 56% |
| Ix | 54% | 69% |
| Iy | 74% | 66% |
| Ow | 48% | 54% |
| Oy | 37% | 39% |
| Uh | 0% | 7% |
| Uw | 21% | 39% |
| Ux | 21% | 63% |
| B | 35% | 48% |
| Bcl | 30% | 40% |
| D | 53% | 51% |
| Dcl | 51% | 44% |
| Dx | 51% | 46% |
| Gcl | 24% | 33% |
| Kc | 42% | 52% |
| P | 28% | 43% |
| Pcl | 14% | 46% |
| Q | 48% | 50% |
| T | 48% | 50% |
| Tcl | 37% | 43% |
| En | 0% | 46% |
| M | 27% | 46% |
| Hh | 34% | 33% |
| Hv | 26% | 35% |
| Jh | 20% | 19% |
| Em | 0% | 0% |
| G | 19% | 46% |
| K | 54% | 56% |
| Ng | 0% | 0% |
| N | 59% | 59% |
| Nx | 5% | 19% |
| Dh | 51% | 50% |
| F | 54% | 48% |
| S | 56% | 51% |
| Sh | 53% | 42% |
| Th | 7% | 6% |
| V | 51% | 44% |
| Zh | 0% | 0% |
| El | 4% | 20% |
| L | 33% | 38% |
| R | 38% | 40% |
| W | 29% | 32% |
| Y | 34% | 33% |
| Epi | 40% | 37% |
| h# | 62% | 57% |
| ch | 21% | 19% |

The table 3 shows that the recognition rates with the proposed hierarchical recognition systems was always superior to the recognition rates with the hierarchical recognition system using phoneme distribution which enhance the inter-phonemes confusions. However, the recognition system find difficult to correctly classify few phonemes with both the two architectures such as em. This problem is related to the so limited number of those sets since they are misrepresented in the speech database.

## 6. Conclusion

In this paper, we have analyzed the phonemes confusions using the speech corpus TIMIT. This confusion analysis shows that many phonemes are confused with other phonemes having similar articulatory structure. Due to this, the robustness of the phoneme recognizer was affected. In an effort to overcome this issue, we designed a new hierarchical phoneme recognition system which takes into account the problematic phonemes causing confusions.

We have shown by this new system that we could improve the phoneme recognition rates when we reduce the inter-phonemes confusions. In future work, we will test whether this conclusion will hold when we use other speech datasets using different dialects and we intend to enhance the robustness of the proposed recognition system by introducing new structure which eliminates completely the phonemes confusions.